\documentclass[lettersize,journal]{IEEEtran}

\usepackage{amsmath,amsfonts}
\usepackage{algorithmic}
\usepackage{algorithm}
\usepackage{array}
\usepackage[caption=false,font=normalsize,labelfont=sf,textfont=sf]{subfig}
\usepackage{textcomp}
\usepackage{stfloats}
\usepackage{url}

\usepackage[T1]{fontenc}

\usepackage[colorlinks=true,
            linkcolor=black,
            filecolor=black,
            urlcolor=black,
            citecolor=black]{hyperref}

\usepackage{verbatim}
\usepackage{graphicx}
\usepackage{cite}
\usepackage{booktabs}
\usepackage{multirow}
\usepackage{lineno}

\begin{document}

\title{DNAlign: Dynamic Null-Space \\ Safe Alignment for LLMs}

\author{Jisheng Dang, Yushuo Zhao, Dewei Liu, Junfeng Fang,  Bimei Wang, Tiantian Rao,\\ Hong Peng,  Bin Hu, ~\IEEEmembership{Fellow, ~IEEE}, Tat-Seng Chua\\

\thanks{Jisheng Dang, Yushuo Zhao, Dewei Liu, Tiantian Rao, Bimei Wang, and Hong Peng are with the School of Information Science and Engineering, Lanzhou University, Lanzhou 730000, China.}

\thanks{Junfeng Fang and Tat-Seng Chua are with the National University of Singapore, Singapore 119077.}

\thanks{Bin Hu is with the School of Medical Technology, Beijing Institute of Technology, Beijing 100081, China}
\thanks{Tat-Seng Chua is with the National University of Singapore, Singapore 119077 (E-mail: dcscts@nus.edu.sg). }
\thanks{This work was supported by the National Natural Science Foundation of China (Grants No. 62227807 and U24B20186). This work was also supported by the Supercomputing Center of Lanzhou University.} 
\thanks{$\ast$Corresponding author
Email addresses: macs@lzu.edu.cn (Bin Hu), wangbm@stu2021.jnu.edu.cn (Bimei Wang), fangjf1997@gmail.com (Junfeng Fang).}
}

\maketitle
\begin{abstract}

Ensuring the safe and reliable deployment of large language models (LLMs) remains a fundamental challenge. Existing safety alignment approaches either incur high computational cost or unintentionally disrupt the model’s core knowledge, leading to degraded fluency and factual accuracy on benign tasks. This reveals a persistent trade-off between safety and utility.
We propose \textbf{DNAlign}, a lightweight alignment framework that integrates control-theoretic optimization with null-space projection. By treating the LLM as a dynamic system, the proposed framework introduces controllable perturbations to steer generation toward safe behavior. A key component is the projection module, which restricts these perturbations to the harmful-related subspace derived from neutral hidden states, thereby preserving general knowledge and response quality. A value function trained on human preference data adaptively optimizes the control signals to align with human safety preferences.
Extensive evaluations across multiple LLM backbones demonstrate that our framework consistently reduces harmful outputs while maintaining fluency, coherence, and factual utility. It achieves superior overall performance compared to prior alignment baselines without sacrificing generation diversity. These results indicate that the proposed framework provides an effective and practically deployable solution for safe LLM alignment. Code is available at \url{https://anonymous.4open.science/r/DNAlign}.
\end{abstract}


\section{Introduction}
\IEEEPARstart{E}{nsuring} the safety of Large Language Models (LLMs) is crucial for their reliable deployment in real-world applications. While LLMs have shown strong performance across fields like healthcare \cite{Rajpurkar2018Cardiologist, Topol2019High}, finance \cite{Chen2023FinGPT}, and education \cite{Adiwardana2020Towards}, they also present significant risks by generating harmful or biased content. For example, chatbots may provide harmful instructions or reinforce negative stereotypes \cite{Bolukbasi2016ManistoComputer}, undermining trust and potentially causing societal harm \cite{Bender2021OnTheDangers, Weidinger2021TruthfulAI}. Thus, two challenges must be addressed: aligning model outputs with human values (safety alignment) and preserving the model’s pre-trained knowledge for factual, fluent responses to benign queries.

\begin{figure}
    \centering
    \includegraphics[width=1\linewidth]{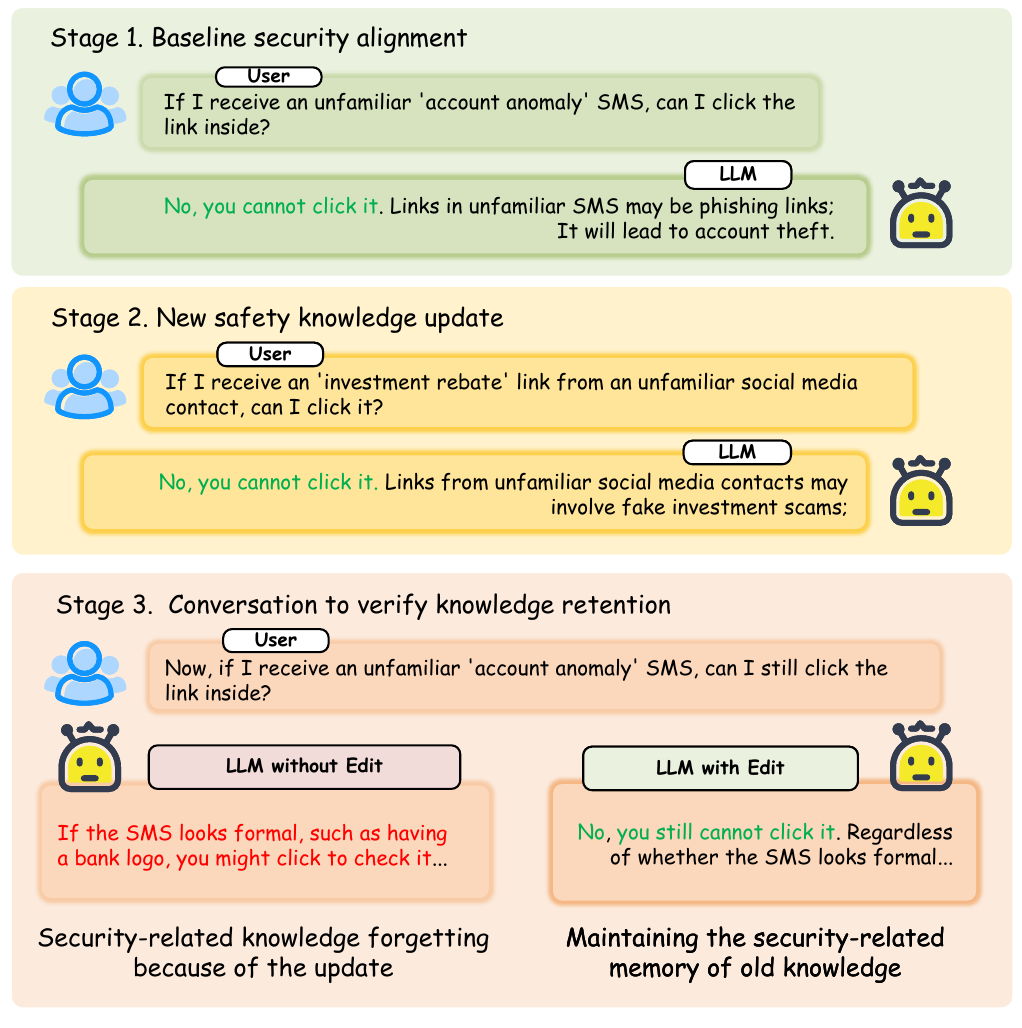}
    \caption{Comparison of LLMs with/without DNAlign alignment. Unedited models forget security-related knowledge, while our proposed framework preserves safe responses.}
    \label{fig:Motivation}
\end{figure}

Achieving robust safety alignment has led to methods like Reinforcement Learning from Human Feedback (RLHF) \cite{kaufmann2024surveyreinforcementlearninghuman, Christiano2017DeepReinforcementLearning} and Direct Preference Optimization (DPO) \cite{Rafailov2023DPO}. While these approaches guide models toward safer outputs, they require extensive human labeling, risk overfitting to specific scenarios \cite{Evans2021TruthfulAI}, and struggle with balancing safety across diverse real-world prompts. Test-time interventions such as prompt engineering \cite{liu2023incremental} and controlled decoding \cite{Kandpal2022DeduplicatingTrainingData, Holtzman2020TheCuriousCase} offer a lightweight alternative, but their robustness remains limited: slight changes in phrasing, domain, or prompt structure often weaken their effect. Existing dynamic intervention methods, although more flexible, can still degrade fluency or factual accuracy on benign tasks, illustrating the difficulty of achieving safety without harming utility.

The second challenge is preserving LLMs’ core knowledge. Safety alignment methods often interfere with models' ability to generate accurate, fluent responses to benign queries. Full fine-tuning \cite{Devlin2019BERT, Radford2018Improving} or static representation editing can overwrite valuable knowledge \cite{Gehman2020RealToxicityPrompts, Mitchell2022Locating}, while continual learning techniques like Elastic Weight Consolidation (EWC) \cite{Kirkpatrick2017OvercomingCatastrophic} do not scale well to large models \cite{liu2023incremental}.
Current efforts to integrate safety alignment and knowledge preservation are fragmented. Post-editing strategies like RLHF after knowledge editing \cite{Wang2023EditingAndAligning} may overwrite edited parameters, and knowledge-aware prompting \cite{Zhou2024FactuallyAligned} improves accuracy but is sensitive to prompt variations and model biases \cite{Dhulia2023LimitsOfPrompting}. These methods lack a unified framework to optimize both safety and knowledge retention in a single process.

We propose DNAlign, a novel LLM safety alignment method that uses control theory and null-space projection to unify safety alignment and knowledge preservation. Our framework models LLMs as dynamic systems, introducing control signals to direct outputs toward safety while ensuring these signals only affect harmful content, preserving pre-trained knowledge. By training a lightweight value function on human preference data, the proposed method optimizes safety alignment while minimizing interference with the model’s core knowledge. These observations highlight the need for a unified mechanism that jointly addresses safety and knowledge preservation, rather than treating them as separate objectives. Our contributions are as follows:

\begin{itemize}
    \item We introduce a unified framework, DNAlign, that unifies efficient safety alignment and knowledge preservation, resolving the trade-off between the two goals.
    \item Grounded in control theory, we design dynamic control signals and null-space constraints to suppress harmful content without interfering with pre-trained knowledge, validated through rigorous experiments.
    \item  We test our method on multiple 7B-scale LLMs (Vicuna-7B and Qwen-7B) and show that it outperforms unedited models and other safety alignment methods in both safety (average reward) and core performance (coherence, diversity), with strong generalization to new harmful scenarios.
\end{itemize}

\section{Related Work}
LLMs have shown remarkable utility across domains, but balancing their safety and core knowledge retention remains a critical unresolved issue. Existing research tends to address these two goals in isolation, leading to trade-offs that limit real-world applicability. This section reviews relevant work from three perspectives, safety alignment, knowledge preservation, and their integration, and clarifies the gaps our framework aims to address.

\subsection{Safety Alignment for LLMs}
To mitigate the inherent risks associated with Large Language Models generating harmful, biased, or unethical content, researchers have developed various alignment methodologies. These approaches can be broadly classified into two categories: pre-training/fine-tuning-based methods which directly modify model parameters, and test-time intervention-based approaches which impose constraints on the output generation process without altering the pre-trained weights.Pre-training and fine-tuning strategies typically leverage extensive human preference data to reshape model behavior. Representative examples include RLHF \cite{Christiano2017DeepReinforcementLearning,kaufmann2024surveyreinforcementlearninghuman} and DPO \cite{Rafailov2023DPO}. RLHF involves training a reward model using human evaluations of model outputs, subsequently fine-tuning the LLM via reinforcement learning to maximize this learned reward signal \cite{Ouyang2022TrainingLanguageModels}. DPO simplifies this pipeline by directly optimizing the model to align with human preference rankings, thereby obviating the need for an explicit reward model. While effective in enhancing safety, these methods face critical limitations: they require substantial, high-quality human labeling, rendering the process resource-intensive and unscalable. 

Moreover, existing methods frequently suffer from overfitting to specific training prompt distributions, resulting in a lack of adaptability to the diverse, unstructured prompts encountered in real-world deployment \cite{Evans2021TruthfulAI}. A more profound concern is their reliance on the LLM’s existing knowledge base, which may lead to compliant but factually inaccurate responses in complex domains \cite{Weidinger2021TruthfulAI}. Test-time interventions offer a lighter alternative by directly adjusting outputs. Techniques such as prompt engineering \cite{liu2023incremental} involve appending safety instrsctions (e.g.,  avoid harmful advice) to the input prompt, while controlled decoding \cite{Kandpal2022DeduplicatingTrainingData,Holtzman2020TheCuriousCase} modifies token sampling (e.g., penalizing harmful tokens) to suppress unsafe outputs. These approaches are straightforward to implement but critically lack robustness. Prompt engineering is highly sensitive to subtle rephrasing, which can easily bypass safety constraints. Controlled decoding, conversely, struggles to adapt to the dynamically varying safety needs across different scenarios (e.g., the differential tolerance standards required for educational versus healthcare contexts). Even sophisticated dynamic intervention methods, which model LLMs as dynamic systems, often fail to strike an adequate balance between safety and utility, frequently resulting in degraded fluency or accuracy on benign tasks \cite{Ganguli2022EmergentAbilities}.

\subsection{Preservation of LLMs’ Core Knowledge}
The ability of LLMs to generate fluent, factual responses to benign queries (e.g., general knowledge retrieval or daily tasks) is fundamentally dependent on the vast knowledge accumulated during their pre-training phase. However, the safety alignment process often interferes with this core knowledge, leading to a noticeable performance degradation in domains unrelated to safety \cite{Gehman2020RealToxicityPrompts,Mitchell2022Locating}. Existing knowledge preservation methods face distinct limitations corresponding to their technical methodologies. Early efforts primarily focused on full-model fine-tuning \cite{Devlin2019BERT,Radford2018Improving}, where all LLM parameters are updated by retraining on new, task-specific data. Although this can partially retain pre-trained knowledge, it is computationally demanding and highly susceptible to catastrophic forgetting, where newly introduced data overwrites previously critical information \cite{Kirkpatrick2017OvercomingCatastrophic}. To mitigate this, continual learning techniques like  EWC \cite{Kirkpatrick2017OvercomingCatastrophic} and memory replay \cite{Rebuffi2017iCaRL} were developed. EWC penalizes updates to parameters deemed critical for old tasks, while memory replay reinforces pre-trained knowledge by resampling and replaying historical task examples. 

Unfortunately, existing solutions are ill-suited for modern LLMs with billions of parameters. EWC requires tracking and storing parameter importance estimates, leading to computationally prohibitive overhead, and memory replay necessitates the storage of massive, potentially sensitive, historical datasets \cite{liu2023incremental}. In contrast, targeted knowledge editing methods offer greater precision by modifying only specific components linked to the knowledge being updated. Some approaches like ROME \cite{Meng2022LocatingAndEditing} and MEMIT \cite{Meng2023MassEditingMemory} employ causal tracing to precisely identify knowledge neurons or attention patterns associated with specific facts (e.g.,  Earth’s orbit) and edit these localized structures to update or correct knowledge. More recently, null-space projection methods \cite{Mitchell2022Locating} have emerged, attempting to protect unrelated knowledge by projecting parameter perturbations into a subspace orthogonal to neutral knowledge representations (e.g., benign prompt features). However, these null-space techniques are predominantly focused on the issue of factual updating and largely neglect behavioral alignment. Consequently, LLMs might retain accurate knowledge but still exhibit harmful behavior, for instance, correctly explaining chemical principles yet providing inappropriate guidance for dangerous experiments.

\subsection{Integration of Safety and Knowledge Preservation}
Despite substantial progress made in the separate domains of safety alignment and knowledge preservation, the effective and unified integration of these two critical goals remains highly fragmented and largely ineffective. Current attempts predominantly fall into two categories: sequential pipelines and prompt-based methods. Sequential approaches, such as performing knowledge editing followed by subsequent RLHF alignment \cite{Wang2023EditingAndAligning, Xu2024ParameterConflicts}, frequently suffer from detrimental parameter interference, where later alignment updates inadvertently overwrite the earlier knowledge corrections, thus failing to maintain factual accuracy. Prompt-based strategies \cite{Zhou2024FactuallyAligned} provide a lightweight, non-intrusive alternative but suffer from extreme brittleness to prompt rephrasings and are fundamentally incapable of addressing the deeper, harmful associations intrinsically embedded within the model's learned representations \cite{Dhulia2023LimitsOfPrompting}. These limitations collectively expose a significant gap in the literature: current methods treat safety and knowledge as distinct, independent objectives rather than optimizing them jointly and holistically within a unified, integrated process.

\section{Preliminary}
To establish a theoretical foundation for DNAlign, this section first elaborates on the generation mechanism of LLMs and then formalizes the mathematical framework for knowledge retention, two core pillars that underpin the design of our safety alignment method.

\subsection{Large Language Models}
LLMs generate fluent and contextually coherent text by predicting the next token based on preceding context, with the Transformer architecture \cite{vaswani2017attention} serving as their core. Simply put, an LLM \( \mathcal{M}(\theta) \) takes a sequence \( x = [x_0, \dots, x_{T-1}] \) and calculates the probability of the next word \( x_T \),

\begin{equation}
p(x_T \mid x_{<T}; \theta) = \text{softmax}\left( \text{LMHead}\left( z_{T-1}(\theta) \right) \right),
\end{equation}
where \( z_{T-1}(\theta) \in \mathbb{R}^d \) is the hidden state from the Transformer’s final layer, driven by parameters \( \theta \), like attention weights \( W^a \) and feed-forward matrices \( W^{ffn} \). The LMHead maps this state to the vocabulary.

The Transformer has \( L \) layers, each turning the input hidden state \( h_{t-1}^{(l-1)} \in \mathbb{R}^d \) into \( h_t^{(l)} \):

\begin{equation}
h_t^{(l)} = \text{LN}\left( h_{t-1}^{(l-1)} + \text{MHA}(W^a) + \text{FFN}(W^{ffn}) \right),
\end{equation}
where \(\text{LN}\) is layer normalization, \(\text{MHA}\) is multi-head attention, and \(\text{FFN}\) is a feed-forward network with activation (like GeLU).

We empirically observe that final-layer hidden states \( z_t(\theta) \) act like high-level “semantic anchors,” encoding prompt meanings and differentiating, for example, “How to bake a cake?” (neutral knowledge) from “How to pick a lock?” (requiring safety handling). Neutral prompt–response pairs \( \mathcal{P}_0 \) represent general knowledge, while harmful pairs \( \mathcal{P}_1 \) require safety alignment. Response quality is scored by a reward function \( r(\text{response} \mid \text{prompt}; \theta) \), reflecting human preferences.

\subsection{Mathematical Framework for Knowledge Retention}
To protect neutral knowledge, we build a framework that ensures prompt–response pairs in \( \mathcal{P}_0 \) are minimally affected by interventions. Let \( K_0 \in \mathbb{R}^{d \times |\mathcal{P}_0|} \) be the matrix of final-layer hidden states for neutral prompts, where each column is the representation \( z_{\text{prompt}_j}(\theta) \) of a neutral prompt. We aim for interventions to keep responses essentially unchanged,

\begin{equation}
r(z_t + u_t \mid \text{prompt}_j) \approx r(z_t \mid \text{prompt}_j),
\end{equation}
where $\quad \forall z_t \in \text{span}(K_0)$ and \( u_t \in \mathbb{R}^d \) is the intervention signal. We therefore constrain \( u_t \) to approximately lie in the null space of \( K_0 \), so that perturbations have minimal impact on neutral knowledge.

\section{Method}
To address the trade-off between large language model safety alignment and core knowledge preservation, we propose \textbf{DNAlign}, a unified framework integrating control theory and null-space projection. This method models LLMs as discrete-time dynamic systems for targeted safety interventions, while leveraging null-space constraints to protect pre-trained neutral knowledge. 

\subsection{Dynamical System Formulation for LLMs}

To address the inherent trade-off between safety alignment and core knowledge preservation in LLMs, we formalize LLMs as discrete-time dynamic systems. This approach is rooted in control theory and explicitly captures the sequential evolution of internal states during token generation. This formalization enables controllable safety guidance by modeling state transitions and control signal interactions, while strictly preserving the LLM’s pre-trained generation logic such as contextual coherence and semantic fluency. Unlike existing safety methods that modify model weights or truncate outputs, which often lead to knowledge degradation, our dynamic system framework restricts interventions to transient state adjustments, ensuring neutral knowledge encoding remains unaltered.

\subsubsection{System State Definition}

The LLM state at time step $ t $ (corresponding to the $ t $-th generated token) is rigorously defined as a tuple of context-encoding and generation-guiding components:

\begin{equation}
s_t = \{h_t, o_t\},
\end{equation}
where each component encodes distinct yet complementary information critical to generation and control.

$ h_t $, a structured collection of attention key–value (KV) pairs from all Transformer encoder–decoder layers (for encoder–decoder LLMs) or decoder layers (for decoder-only LLMs). Each KV pair $ (k_{l,t}, v_{l,t}) $ in $ h_t $ (where $ l \in [1, L] $, $ L $ is the number of layers) encodes contextual information accumulated from the input prompt and previously generated tokens. Specifically, $ k_{l,t} $ captures token-level semantic features, while $ v_{l,t} $ stores context-dependent relevance weights. Together, $ h_t $ represents the LLM’s contextual memory that governs coherent multi-turn generation.
$ o_t \in \mathbb{R}^d $, the pre-logit activation vector from the final Transformer layer, immediately before the language model head, LMHead. This vector directly parameterizes the next token’s probability distribution via $ \text{softmax}(W o_t) $ (where $ W \in \mathbb{R}^{V \times d} $, $ V $ is the vocabulary size), making $ o_t $ the primary target for fine-grained generation control.

The initial state $ s_0 = \{h_0, o_0\} $ is derived by feeding the full input prompt (such as a user query or multi-turn dialogue history) through the LLM’s Transformer layers. Notably, $ h_0 $ and $ o_0 $ are computed using the LLM’s original pre-trained forward pass. This ensures the initial contextual encoding aligns with the model’s native knowledge representation, which is critical for preserving neutral knowledge.

\begin{figure*}[t]
  \centering
  \includegraphics[width=1\textwidth, height=8cm, keepaspectratio]{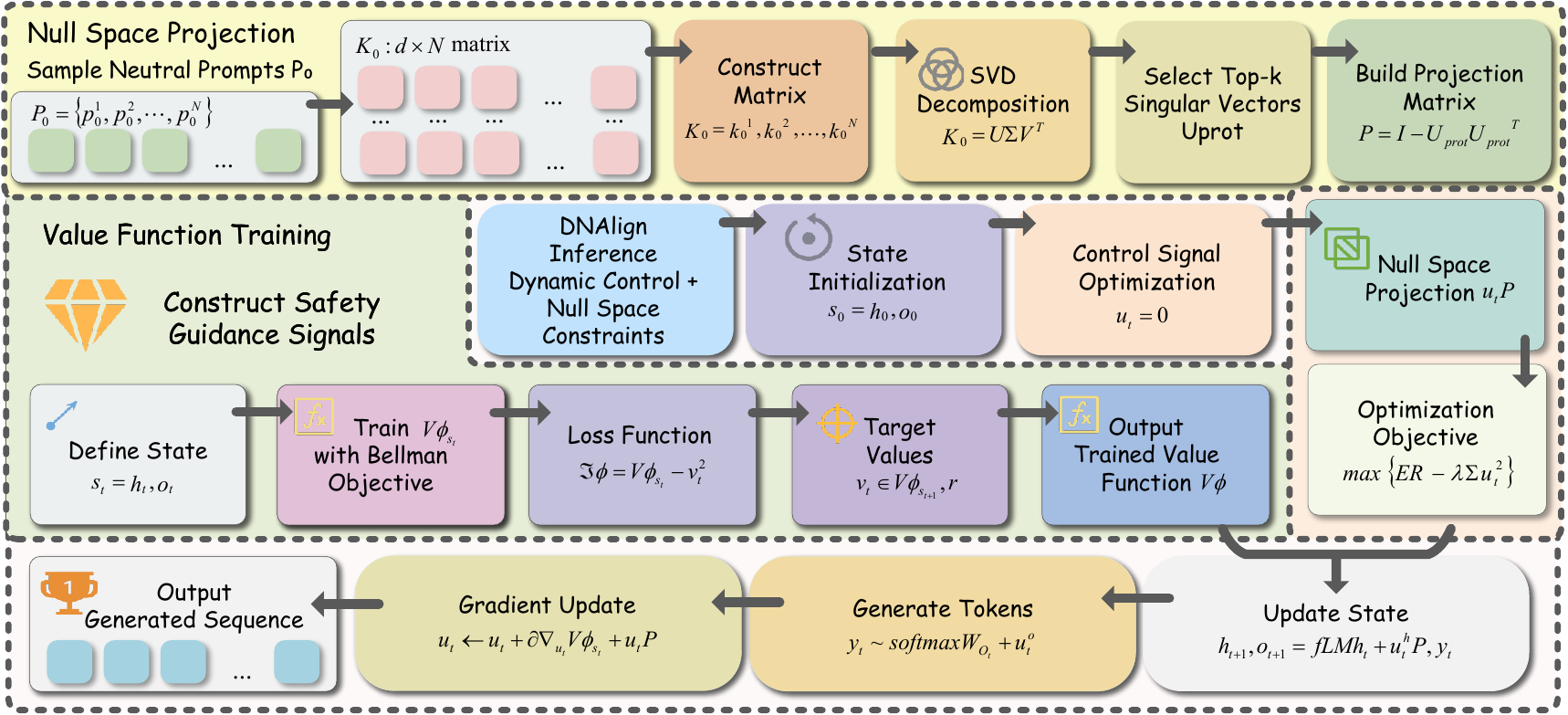}
  \caption{The sequential process begins by computing the initial state $s_0=\{h_0, o_0\}$ from the input prompt. It then loads the pre-computed null-space projection matrix $P$. In the core step, a gradient-optimized control signal $u_t$ is projected by $P$ to enforce safety alignment while minimally affecting neutral knowledge. Finally, the projected signal $P u_t$ updates the system state $s_{t+1}$, enabling token generation $y_t$.}
  \label{fig:method_overview}
\end{figure*}

\subsubsection{State Transition and Control Signal Injection}

In the absence of safety interventions, the LLM’s state transitions follow its inherent, pre-trained mechanism. This mechanism is governed by Transformer self-attention and Feed-Forward Network (FFN) operations:

\begin{equation}
h_{t+1}, o_{t+1} = f_{\text{LM}}\left(h_t, y_t\right),
\end{equation}
where $ f_{\text{LM}} $ denotes the LLM’s native state transition function. This function updates $ h_{t+1} $ by integrating the $ t $-th generated token $ y_t $ into the existing KV context via attention weight recalibration and computes $ o_{t+1} $ via the final layer’s FFN. The token $ y_t $ is sampled from the probability distribution $ y_t \sim \text{softmax}\left(W o_t\right) $, where $ W $ is the LMHead’s weight matrix. This matrix is fixed during inference to preserve pre-trained generation quality.

To suppress harmful outputs such as instructions for dangerous activities or misleading information without altering neutral generation, we inject dynamic control signals $ u_t = \{u_t^h, u_t^o\} $ into the state transition process. These signals are task-specific and transient in the sense that they do not modify model parameters and are re-computed at each step:

\begin{equation}
\begin{cases}
y_t \sim \text{softmax}\left(W\bigl(o_t + u_t^o\bigr)\right),\\[2pt]
h_{t+1},\, o_{t+1} = f_{\text{LM}}\left(h_t + u_t^h, y_t\right),
\end{cases}
\end{equation}
where $ u_t^o \in \mathbb{R}^d $ adjusts the pre-logit vector to bias the token distribution away from harmful candidates, such as reducing the probability of tokens like \emph{explosive} in safety-critical queries, while $ u_t^h $ (a structured KV adjustment matching the dimension of $ h_t $) ensures the updated context $ h_t + u_t^h $ remains semantically consistent, avoiding nonsensical transitions.

\subsection{Null-Space Constraints for Knowledge Preservation}

A critical challenge in LLM safety alignment is avoiding unintended disruptions to pre-trained neutral knowledge. Such knowledge underpins fluent and factual responses to benign queries, and modifications to it often lead to semantic incoherence or factual errors in non-harmful generation scenarios. To address this, we introduce null-space projection constraints. These constraints explicitly restrict the action scope of safety-oriented control signals to subspaces that are mathematically orthogonal to the subspaces encoding neutral knowledge, ensuring that safety interventions only target harmful content without altering the LLM’s native ability to process neutral prompts.

\subsubsection{Construction of the Projection Matrix}

The core of our null-space constraint mechanism lies in constructing a projection matrix that delineates the boundary between neutral knowledge subspaces and harmful content subspaces. To this end, we first extract high-dimensional hidden state representations of neutral prompts from the SHP dataset. This dataset is selected for its diverse coverage of benign scenarios, including daily conversations, factual queries, and general instruction-following tasks, which ensures the extracted neutral knowledge is representative of real-world non-harmful use cases.
Let $ P_0 = \{p_0^1, p_0^2, \dots, p_0^N\} $ denote a set of $ N $ neutral prompts sampled from the SHP dataset. For each prompt $ p_0^i $, we feed it into the pre-trained LLM and extract the final-layer hidden state (consistent with the $ z_t $ definition) after full contextual encoding, denoted as $ k_0^i \in \mathbb{R}^d $. We then aggregate these individual hidden states into a matrix $ K_0 \in \mathbb{R}^{d \times N} $, where each column corresponds to the hidden state of one neutral prompt in $ P_0 $.

To define the  protected subspace that encodes neutral knowledge, we perform singular value decomposition (SVD) on $ K_0 $ to capture its principal components. Let $ K_0 = U \Sigma V^T $ be the SVD result, where $ U \in \mathbb{R}^{d \times d} $ is the matrix of left singular vectors, $ \Sigma \in \mathbb{R}^{d \times N} $ is the diagonal matrix of singular values, and $ V \in \mathbb{R}^{N \times N} $ is the matrix of right singular vectors. We retain the first $ k $ left singular vectors (corresponding to the top $ k $ largest singular values) to form the basis matrix $ U_{\text{prot}} \in \mathbb{R}^{d \times k} $ of the protected subspace. The value of $ k $ is determined via cross validation to balance the completeness of neutral knowledge coverage and the compactness of the subspace.

Finally, the projection matrix $ P \in \mathbb{R}^{d \times d} $ that maps signals into the null space of the protected subspace is computed as:

\begin{equation}
P = I_d - U_{\text{prot}} U_{\text{prot}}^T,
\end{equation}
where $ I_d \in \mathbb{R}^{d \times d} $ is the identity matrix. This formulation ensures that any vector multiplied by $ P $ is orthogonal to the protected subspace, as $ U_{\text{prot}}^T P = 0 $ by construction.

\subsubsection{Constraining Control Signals}

The projection matrix $ P $ is integrated into the control signal injection process to enforce null-space constraints. Specifically, before injecting the control signals $ u_t = \{u_t^h, u_t^o\} $ into the LLM’s state transition, we project both components onto the null space of the protected subspace using $ P $. This modified injection process is formalized as:

\begin{equation}
h_{t+1}, o_{t+1} = f_{\text{LM}}\left(h_t + u_t^h P, y_t\right),
\end{equation}
where $ u_t^h P $ denotes the projection of the KV-pair control component $ u_t^h $ (with dimension matching $ h_t $), and the projection of $ u_t^o $ (applied to the pre-logit vector) follows the same logic (omitted here for notational consistency).

This projection step guarantees two key properties. First, the modified control signals $ u_t^h P $ and $ u_t^o P $ have no overlapping components with the protected subspace encoding neutral knowledge. This means they cannot alter the LLM’s internal representations of neutral content, theoretically preserving the model’s ability to generate fluent and factual responses to benign prompts. Second, the projected signals retain their ability to act on subspaces associated with harmful content, as these subspaces are orthogonal to the protected subspace by design. This ensures the safety alignment objective of suppressing harmful outputs remains uncompromised.

Notably, the projection matrix $ P $ is computed offline using neutral prompt data. This design avoids introducing additional computational overhead during online inference, as the matrix multiplication operation $ u_t P $ is computationally lightweight compared to the LLM’s native Transformer operations.

\subsection{Control Signal Optimization}
We optimize control signals to maximize safety alignment with human preferences while minimizing perturbations to maintain generation quality, using a value function trained on human preference data.

\begin{figure*}[t]
\centering
\includegraphics[width=0.99\textwidth, height=10cm, keepaspectratio]{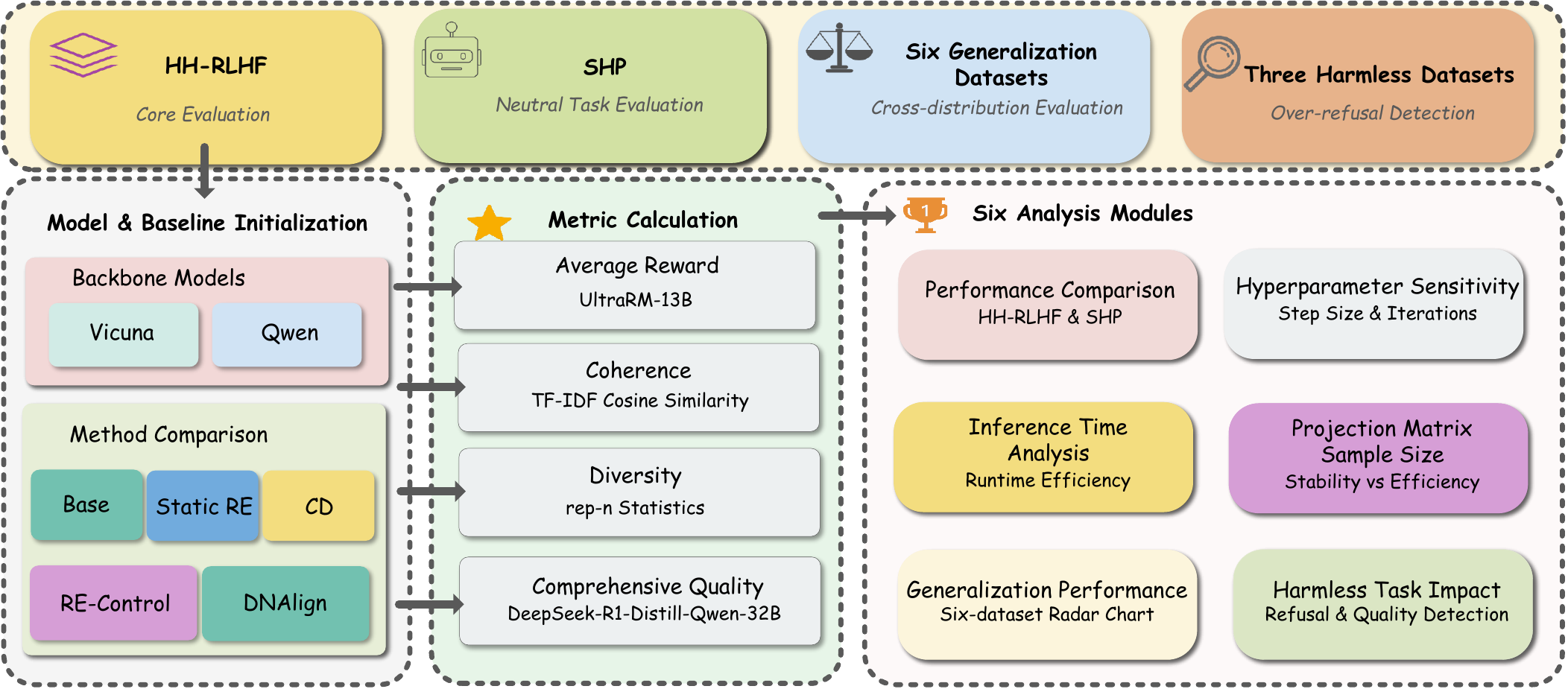}
\caption{End-to-end experimental workflow of the proposed DNAlign framework. This workflow includes four key stages: (1) data preparation (curating neutral prompts from the SHP dataset and harmful prompts from benchmark datasets), (2) offline projection matrix construction (extracting hidden states of neutral prompts to define protected subspaces), (3) online control signal injection (projecting signals into null spaces before modifying LLM states), and (4) performance evaluation (assessing both safety alignment and neutral knowledge preservation via multi-metric validation).}

\label{fig:exp_workflow}
\end{figure*}

\subsubsection{Value Function Training}
We train a value function \( V_\phi(s_t) \) (a 3-layer MLP parameterized by \( \phi \)) to estimate the long-term reward of state \( s_t \), reflecting how well the state leads to safe, human-preferred outputs. The value function is trained on the HH-RLHF dataset (human-labeled safe/helpful pairs) by minimizing the Bellman loss:  

\begin{equation}
\mathcal{L}(\phi) = \sum_{i} \sum_{m} \sum_{t}
\left(
  V_{\phi}\left(s_{t}^{i, m}\right)
  - \text{stop-grad}\left(v_{t}^{i, m}\right)
\right)^{2},
\end{equation}
where \( s_t^{i,m} \) is the \( t \)-step state of the \( m \)-th sample in the \( i \)-th batch, \( v_t^{i,m} \) is the target value (either \( V_\phi(s_{t+1}^{i,m}) \) or the final reward \( r^{i,m} \) from UltraRM-13B, a reward model prioritizing harmlessness), and \( \text{stop-grad} \) prevents gradient propagation to next-state estimates.

\subsubsection{Gradient Ascent for Control Signal Update}
During inference, we optimize \( u_t \) via gradient ascent to maximize expected rewards while limiting perturbations:  

\begin{equation}
\underset{\{u_t\}}{\arg\max}\;
\mathbb{E}[R] - \lambda \sum_{t=1}^{T}\left\|u_{t}\right\|_2^{2},
\end{equation}
where \( \mathbb{E}[R] \) is the expected long-term reward, and \( \lambda \) is the regularization coefficient for perturbation control.  

The update rule for \( u_t \) is:  

\begin{equation}
u_{t} \leftarrow u_{t} + \alpha \nabla_{u_{t}} V_{\phi}\bigl(s_{t} + u_{t} P\bigr),
\end{equation}
where \( \alpha \) is the step size, and we perform 100 iterations of gradient ascent to refine \( u_t \), ensuring it suppresses harm without introducing incoherence.

\subsection{End-to-End Workflow of DNAlign}

To unify safety alignment and knowledge preservation in LLM inference, DNAlign employs a four-step sequential end-to-end workflow. This design aligns with our experimental framework (Section V) to ensure reproducibility and minimal computational overhead, critical for efficient real-world deployment.

\textbf{Initial State Computation}. The input prompt (single/multi-turn) is fed into the pre-trained LLM (e.g., Vicuna-7B) to compute the initial state \( s_0 = \{h_0, o_0\} \), where \( h_0 \) denotes decoder-layer attention key–value (KV) pairs and \( o_0 \) is the final-layer pre-logit vector. Using unmodified pre-trained parameters ensures \( s_0 \) faithfully encodes the model’s native understanding of both neutral and harmful content, preserving core knowledge before safety interventions.

\textbf{Projection Matrix Loading}. The null-space projection matrix \( P \), enforcing orthogonality between control signals and neutral knowledge subspaces, is pre-computed offline. Via singular value decomposition (SVD) of the SHP dataset-derived neutral knowledge matrix \( K_0 \), we calculate \( P = I_d - U_{\text{prot}} U_{\text{prot}}^\top \) (Section IV.B). Loading pre-computed \( P \) at inference avoids on-the-fly decomposition, consuming $< 0.1\%$ of total inference time on Vicuna-7B and enabling low-latency deployment.

\textbf{Control Signal Optimization}. This core step adapts safety alignment via gradient-based optimization and null-space projection. For each token step \( t \), a lightweight 3-layer MLP value function \( V_\phi \) (trained on HH-RLHF data) estimates the reward of state \( s_t \), guiding gradient ascent (Eq. 10) to optimize control signal \( u_t = \{u_t^h, u_t^o\} \). The optimization maximizes safety-aligned reward while minimizing \( \|u_t\|_2 \) to avoid over-perturbation. \( u_t \) is then projected via \( P \) to eliminate overlap with protected subspaces, preserving neutral knowledge.

\textbf{State Transition and Token Generation}. The projected signal \( P u_t \) updates the system state to \( s_{t+1} = \{h_t + P u_t^h, o_t + P u_t^o\} \), adjusting KVs for coherence and pre-logits to suppress harmful tokens. The next token \( y_t \) is sampled from \( \text{softmax}(W(o_t + P u_t^o)) \), where \( W \) denotes fixed LMHead weights. This process repeats until an EOS token is generated or the 200-token limit (consistent across experiments) is reached. This workflow avoids full-model fine-tuning (a key limitation of RLHF) via offline \( P \) preprocessing and lightweight \( V_\phi \). We demonstrate in Section~\ref{sec:experiments} that this design achieves a favorable balance between safety alignment and knowledge preservation.

\section{Experiments}
\label{sec:experiments}
The efficacy and robustness of the proposed DNAlign methodology are rigorously evaluated through a set of comprehensive experiments designed to address four core research questions ($\text{RQ}$s):

\noindent\textbf{RQ1 (Safety Alignment).} Does the DNAlign intervention effectively reduce the generation of harmful, toxic, or unethical content, achieving superior safety performance compared to state-of-the-art baselines?

\noindent\textbf{RQ2 (Utility Preservation).} Can DNAlign successfully preserve the LLM's core knowledge and maintain high output quality and fluency when performing harmless or neutral tasks, thereby avoiding the common safety-utility trade-off?

\noindent\textbf{RQ3 (Generalization Capability).} Does the method demonstrate strong generalization to diverse and novel harmful scenarios, specifically those beyond the specific datasets used for constructing the projection space and training the initial value function?

\noindent\textbf{RQ4 (Practical Deployment Efficiency).} Is DNAlign computationally efficient enough for practical, real-world deployment, and how do the key hyperparameters (e.g., step size, number of iterations) influence the resulting safety and utility performance metrics?

To validate the efficacy of our method in achieving a crucial balance between stringent safety alignment and the preservation of core LLM knowledge, we perform comprehensive experiments leveraging the detailed design specifications and data from referenced materials. The subsequent sections systematically elaborate on the research workflow, precise experimental setup, and configurations of established baselines. Following this, we present detailed analyses focusing on six critical experimental aspects. Firstly, performance comparison against established safety baselines. Secondly hyperparameter sensitivity studies (focusing on step size and gradient ascent iterations). Thirdly inference time and computational overhead analysis. Fourthly ablation studies investigating the impact of projection-matrix sample size. Rigorous generalization testing across external harmful benchmarks. Last, detailed impact analysis on a diverse suite of harmless datasets. These structured evaluations provide empirical evidence supporting the method's superiority and practical viability.

\subsection{Research Workflow}
The experiment follows a four-stage sequential workflow to ensure systematic validation: 

\textbf{1) Preparatory Stage}. We construct the null-space projection matrix using neutral prompts from SHP and train the value function on HH-RLHF under reward-model supervision.

\textbf{2) Initialization Stage}. All backbone models (Vicuna-7B and Qwen-7B) and baseline methods are configured with unified random seeds and hyperparameters to ensure fair and consistent comparison.

\textbf{3) Inference and Data Collection Stage}. Our method and all baseline methods are applied to multiple datasets including HH-RLHF, SHP, harmful benchmarks, and harmless benchmarks, to generate outputs and compute the four core evaluation metrics.

\textbf{4) Result Analysis Stage}. Performance across datasets is aggregated to examine the safety–utility trade-off and to support extended studies such as hyperparameter sensitivity, inference-time evaluation, and projection-matrix ablations.

\subsection{Experimental Setup}

\textbf{Model}. We evaluate DNAlign on two base models, Qwen-7B \cite{qwen-7b} and Vicuna-7B \cite{zheng2023judging}. We compare DNAlign with several baselines, including the unedited backbone models, Static RE \cite{li2023inference}, Contrastive Decoding (CD) \cite{khanov2024args}, and RE-Control \cite{kong2024aligning}, which represent methods with static or dynamic safety interventions.

\textbf{Datasets}. Experiments are conducted on multiple datasets covering both harmful and benign scenarios. HH-RLHF and SHP are used as mixed benchmarks to evaluate the trade-off between safety alignment and capability preservation under different data distributions. For harmful generalization, we use RealToxicityPrompts, ToxiGen, and Categorical-HarmfulQA. To assess performance under benign conditions, we evaluate on six harmless datasets: OSASST1, Dolly, HC3, Alpaca, ShareGPT, and Reddit ELI5.

\textbf{Baselines}. The baselines include: (1) Unedited Backbones without safety intervention; (2) Static RE \cite{li2023inference}, a static reward-based activation adjustment method; (3) Contrastive Decoding (CD) \cite{khanov2024args}, which dynamically integrates reward signals during generation; and (4) RE-Control \cite{kong2024aligning}, a dynamic control-based alignment approach.

\subsection{Evaluation Metrics and Tools}
To systematically assess the performance of DNAlign in balancing LLM safety alignment and core knowledge preservation, we adopt four well-established metrics that jointly cover safety-oriented (RQ1), utility-oriented (RQ2), and practicality-related aspects (RQ3 and RQ4). This selection aligns with prior work in LLM alignment and knowledge retention \cite{khanov2024args,brown2020language}, ensuring that our evaluation results are comparable to existing state-of-the-art methods and grounded in community-recognized standards. Each metric is designed to target a specific aspect of model performance, with clear computational definitions and reliable tools to avoid subjective biases, thus enabling rigorous quantitative analysis of the proposed method’s effectiveness.

\textbf{Average Reward}. The average reward metric measures model alignment with human preferences, focusing on safety, helpfulness, and appropriateness core goals of LLM safety alignment. Calculated using UltraRM-13B \cite{cui2023ultrafeedback}, a reward model trained on large-scale human preference data, it quantifies how well model outputs align with these goals. The formula is:
\begin{equation}
\bar{R} = \frac{1}{M} \sum_{i=1}^M r_i,\quad r_i = \text{UltraRM-13B}(\text{inputs}_i),    
\end{equation}
where $ M $ is the total number of test samples (1000), $ \text{inputs}_i $ is the model’s response to the $ i $-th prompt, and $ r_i $ is the unnormalized score from UltraRM-13B. Higher $ \bar{R} $ values indicate better alignment with human preferences for safe and helpful interactions, reflecting DNAlign’s ability to optimize safety without compromising utility.

\textbf{Coherence}. The coherence metric measures the semantic similarity between the model’s generated responses and human-preferred reference responses, indicating core knowledge preservation. By assessing how well the model retains the semantic content of high-quality reference outputs, it ensures safety interventions do not distort the model's understanding of neutral knowledge. Coherence is computed using TF-IDF cosine similarity, a method robust to wording variations while capturing core meaning. The formula is:

\begin{equation}
\text{Coherence} = \frac{\boldsymbol{e}_i \cdot \boldsymbol{r}_i}{|\boldsymbol{e}_i| |\boldsymbol{r}_i|},    
\end{equation}
where $ \boldsymbol{e}_i $ is the TF-IDF vector of the model’s response for the $ i $-th prompt, and $ \boldsymbol{r}_i $ is the TF-IDF vector of the human-preferred reference. Both vectors are projected into a 1000-dimensional space, based on the top 1000 frequent terms from the reference corpus. Higher coherence values indicate that the model’s output is semantically close to the reference, ensuring that DNAlign’s safety interventions do not compromise factual accuracy or contextual relevance.

\textbf{Diversity}. The diversity metric evaluates the variety in the model’s generated responses, ensuring that safety interventions do not lead to repetitive or overly constrained outputs. We measure diversity using rep-$n$ scores, which quantify the repetition rate of $n$-grams (2-grams, 3-grams, and 4-grams) in the generated text. Repeated $n$-grams can make responses feel unnatural, even if they are safe and factual. The formula for diversity is:

\begin{equation}
\text{Diversity} = \prod_{n=2}^{4} \left(1 - \frac{\text{rep-}n}{100}\right),
\end{equation}
where rep-$n$ is the percentage of repeated $n$-grams in the generated response. For each response, we split the text into token sequences, count n-gram frequencies, and compute the proportion of $n$-grams with frequency $\geq$ 2. The final score reflects repetition across 2-gram to 4-gram phrase lengths, with higher values indicating richer, more varied language. Higher diversity scores confirm that DNAlign’s safety mechanisms do not constrain the model's generation, maintaining natural, engaging responses.

\setlength{\tabcolsep}{0.00001pt}
\begin{table*}[htbp]
  \centering
  \caption{Comprehensive performance comparison on the HH-RLHF and SHP datasets.}
  \renewcommand{\arraystretch}{1.12} 
  \setlength{\tabcolsep}{5pt}        
  \begin{tabular}{c c c c c c c c c c c}
    \toprule
    \textbf{Backbone} & \textbf{Dataset} & \textbf{Model} &
    \textbf{Diversity} &  & \textbf{Coherence} &  &
    \textbf{Average Reward} &  & \textbf{Comprehensive Quality} &  \\
    \midrule

    \multirow{10}{*}{Vicuna-7B \cite{zheng2023judging}}
      & \multirow{5}{*}{HH-RLHF}
      & Base
        & 0.8672 & -- & 0.3991 & -- & 0.1410 & -- & 5.3600 & -- \\
      & & CD \cite{khanov2024args}
        & 0.8810 & –  & 0.4081 & –  & 0.8440 & ↑  & 5.6800 & ↑  \\
      & & Static RE \cite{li2023inference}
        & 0.8941 & ↑  & 0.3813 & –  & 0.9130 & ↑  & 5.0500 & ↓  \\
      & & RE-Control \cite{kong2024aligning}
        & 0.9007 & ↑  & 0.3887 & –  & 0.9614 & ↑  & 5.9400 & ↑  \\
      & & \textbf{DNAlign (Ours)}
        & \textbf{0.9026} & \textbf{↑} & \textbf{0.4325} & \textbf{↑}
        & \textbf{1.2258} & \textbf{↑} & \textbf{6.1600} & \textbf{↑} \\
      \cmidrule(lr){2-11}

      & \multirow{5}{*}{SHP}
      & Base
        & 0.8833 & -- & 0.3762 & -- & 0.3913 & -- & 6.3300 & -- \\
      & & CD \cite{khanov2024args}
        & 0.8778 & –  & 0.3809 & –  & 0.7830 & ↑  & 6.2300 & ↓  \\
      & & Static RE \cite{li2023inference}
        & 0.8809 & –  & 0.3792 & –  & 0.7831 & ↑  & 5.5250 & ↓  \\
      & & RE-Control \cite{kong2024aligning}
        & 0.8820 & –  & 0.3792 & –  & 0.7456 & ↑  & 6.5000 & ↑  \\
      & & \textbf{DNAlign (Ours)}
        & \textbf{0.8818} & \textbf{–} & \textbf{0.3813} & \textbf{–}
        & \textbf{1.0756} & \textbf{↑} & \textbf{6.9700} & \textbf{↑} \\

    \midrule

    \multirow{10}{*}{Qwen-7B \cite{qwen-7b}}
      & \multirow{5}{*}{HH-RLHF}
      & Base
        & 0.9381 & -- & 0.4192 & -- & 0.5519 & -- & 7.1538 & -- \\
      & & CD \cite{khanov2024args}
        & 0.8680 & ↓  & 0.3929 & –  & 0.9036 & ↑  & 6.2000 & ↓  \\
      & & Static RE \cite{khanov2024args}
        & 0.9350 & –  & 0.4168 & –  & 0.7088 & ↑  & 6.5750 & ↓  \\
      & & RE-Control \cite{kong2024aligning}
        & 0.9364 & –  & 0.4203 & –  & 0.7218 & ↑  & 7.2500 & ↑  \\
      & & \textbf{DNAlign (Ours)}
        & \textbf{0.9371} & \textbf{–} & \textbf{0.4250} & \textbf{↑}
        & \textbf{0.9430} & \textbf{↑} & \textbf{7.3846} & \textbf{↑} \\
      \cmidrule(lr){2-11}

      & \multirow{5}{*}{SHP}
      & Base
        & 0.9396 & -- & 0.4126 & -- & 0.3939 & -- & 6.6551 & -- \\
      & & CD \cite{khanov2024args}
        & 0.9199 & ↓  & 0.4247 & ↑  & 0.6415 & ↑  & 6.7710 & ↑  \\
      & & Static RE \cite{li2023inference}
        & 0.9338 & –  & 0.4179 & –  & 0.6546 & ↑  & 6.0300 & ↓  \\
      & & RE-Control \cite{kong2024aligning}
        & 0.9424 & ↑  & 0.4148 & –  & 0.6401 & ↑  & 6.6900 & ↑  \\
      & & \textbf{DNAlign (Ours)}
        & \textbf{0.9402} & \textbf{–} & \textbf{0.4164} & \textbf{–}
        & \textbf{0.9882} & \textbf{↑} & \textbf{6.8965} & \textbf{↑} \\
    \bottomrule
  \end{tabular}
\label{tab:model_performance}
\end{table*}

\textbf{Comprehensive Quality Score}. The comprehensive quality score provides an overall assessment of model performance by integrating multiple aspects, including helpfulness, accuracy, fluency, and safety. This metric complements the individual metrics by capturing trade-offs between different performance dimensions, such as whether improved safety affects fluency. It uses DeepSeek-R1-Distill-Qwen-32B \cite{guo2025deepseek}, a state-of-the-art evaluation model validated on diverse LLM benchmarks, to assign scores between 1 and 10 to each response. A score of 1 indicates poor quality (e.g., harmful or inaccurate), while 10 reflects exceptional quality (e.g., safe, accurate, fluent). DeepSeek-R1-Distill-Qwen-32B has a 0.82 correlation with human evaluator scores, making it a reliable tool for capturing nuanced quality differences. Together with the other metrics, this provides a rigorous, objective evaluation framework for assessing DNAlign’s safety alignment and knowledge preservation.

\subsection{Performance Comparison on HH-RLHF and SHP}

The efficacy of our proposed methodology is comprehensively validated across two distinct yet complementary human-preference benchmarks: HH-RLHF and SHP. This dual-benchmark strategy allows for a robust assessment of two critical research objectives, including RQ1, focusing on safety alignment, and RQ2, pertaining to capability preservation in general-purpose interactions. The HH-RLHF dataset, characterized by its mixture of harmful and benign prompts, serves as a crucial testbed for evaluating safety-oriented performance. 

Our proposed approach consistently demonstrates superior safety alignment compared to the RE-Control baseline and other tested systems, across both the Vicuna-7B and Qwen-7B foundational models. Specifically, with the Vicuna-7B backbone, our method achieves a substantially higher average reward (1.2258 vs. 0.9614 for RE-Control), alongside notably improved coherence (0.4325 vs. 0.3887), and better comprehensive quality (6.16 vs. 5.94), while maintaining comparable diversity. A similar positive pattern is evident under the Qwen-7B backbone, where our method significantly increases the average reward from 0.7218 to 0.9430 and registers the highest comprehensive quality. These compelling quantitative results collectively demonstrate that a more stringent safety alignment (RQ1) can be effectively realized without compromising the fluency or content richness of the generated responses in mixed harmful–benign scenarios.

To rigorously verify capability preservation (RQ) under general-purpose, non-adversarial conditions, we employ the SHP dataset, which predominantly consists of harmless, everyday human-model queries. In this context, our method successfully maintains or actively enhances the core model's utility. For the Vicuna-7B model, both the average reward (1.0756) and comprehensive quality (6.97) metrics exceed those of the RE-Control baseline, with diversity and coherence remaining stable. 

The experimental results under the Qwen-7B backbone exhibit the same positive pattern, definitively confirming that the introduced intervention strategy does not erode general-purpose performance. Taken together, the consistent performance improvements across these two structurally different datasets underscore that the proposed approach offers a balanced and synergistic enhancement. It simultaneously elevates both safety-oriented and utility-oriented metrics while diligently preserving broad output diversity. This demonstrated robustness, spanning performance and generalizability, suggests that the method is highly suitable for real-world deployments where the simultaneous requirement for uncompromising safety and high-fidelity response quality is a necessity.

\begin{figure*}
\centering
\includegraphics[width=1\textwidth]{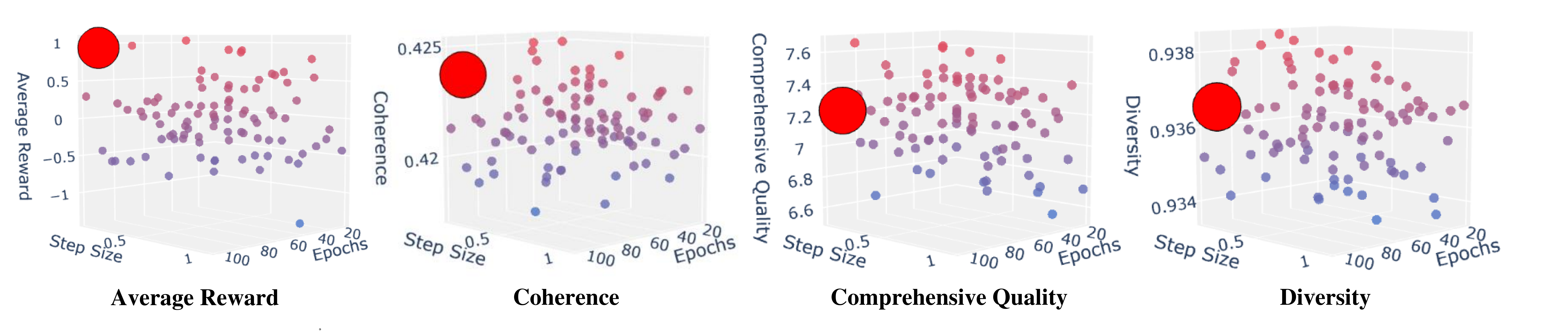}
\caption{The influence of step size and epochs at test-time.}
\label{fig:hyper}
\end{figure*}

\subsection{Hyperparameter Sensitivity Study}
To validate the practicality and robustness of our method, we conduct an ablation study analyzing its sensitivity to key hyperparameters.
To rigorously assess this, we performed systematic experiments utilizing the Qwen-7B model architecture trained on a distilled subset of the HH-RLHF dataset, comprising 1000 representative samples, while systematically varying two critical training parameters: the number of training epochs ($\tau \in [10, 100]$) and the gradient step size ($\eta \in [0.1, 1.0]$). The impact of these combinations was quantitatively evaluated based on four primary metrics,including diversity, coherence, comprehensive quality, and average reward. The detailed results (visualized in Figure 4) reveal three distinct trends. First, Diversity remains notably stable, exhibiting a narrow variance range ($0.933$ to $0.942$) unless epochs are below $\tau=30$ or the step size exceeds $\eta=0.8$. Second, Coherence is the most sensitive metric, achieving peak performance with $\tau \in [40, 70]$ epochs and a moderate step size of $\eta \in [0.2, 0.5]$, but suffering a sharp degradation under insufficient training or overly aggressive policy updates. Third, both Comprehensive Quality and the Average Reward improve with increased training epochs but consistently plateau or begin to decline when the step size is increased beyond $\eta=0.5$. 

Based on this granular analysis, the optimal configuration is identified as $\mathbf{100}$ epochs ($\tau=100$) and a controlled step size of $\mathbf{0.2}$ ($\eta=0.2$), yielding a highly balanced performance profile with a diversity of $0.9366$, coherence of $0.4238$, comprehensive quality of $7.24$, and the highest observed average reward ($\mathbb{R}_{\text{avg}}=0.943$), thereby ensuring both the reliability and practicality of our proposed method.

\begin{figure}
    \centering
    \includegraphics[width=1\linewidth]{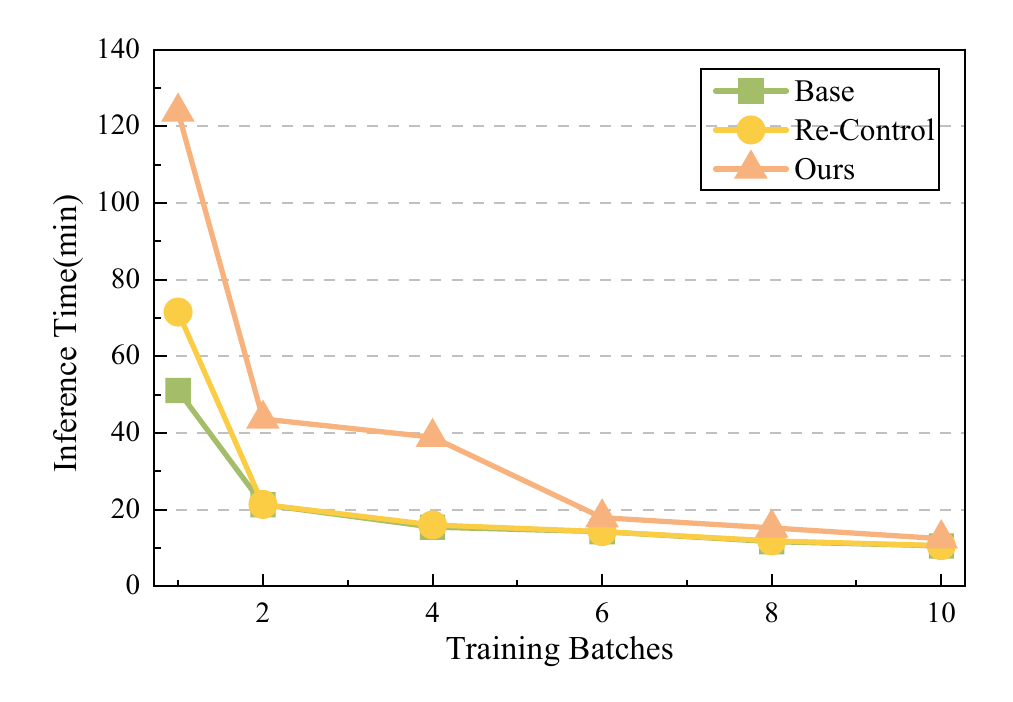}
    \caption{Variation of inference time across three models with training batches.
}
    \label{fig:placeholder}
\end{figure}

\subsection{Inference Time Analysis}
Beyond effectiveness and hyperparameter robustness, we further quantify the computational overhead introduced by our method to evaluate its deployment practicality (RQ4).
To assess the practical deployment of the proposed framework, we evaluate inference time using the Qwen-7B backbone with 1,000 samples from the HH-RLHF dataset.  The experiment compares DNAlign against the RE-Control model and the unedited base model, analyzing the impact of training batches (ranging from 1 to 10) on inference time.

Results show that the base model’s inference time decreases as batch size increases, from 51.059 minutes with 1 batch to 10.556 minutes with 10 batches.  The RE-Control model exhibits a similar trend, dropping from 71.518 minutes (1 batch) to 10.617 minutes (10 batches).  DNAlign, initially slower (123.734 minutes with 1 batch), also improves as batch size increases, reaching 12.447 minutes with 10 batches, only 1.891 minutes longer than the base model and 1.83 minutes longer than RE-Control.This higher initial overhead is due to DNAlign's null-space projection module, critical for safety and knowledge preservation.  However, the inference time gap narrows significantly with more batches, with DNAlign showing improved efficiency in larger batch sizes due to batch processing optimization.  The precomputed projection matrix introduces negligible runtime overhead (less than 0.1\% of total inference time), as it is loaded during inference rather than computed on the fly.
Importantly, our method maintains high output quality despite the additional inference time.  It achieves a coherence score of 0.4250, surpassing both the base model (0.4192) and the RE-Control model (0.4203). This demonstrates that DNAlign successfully balances safety alignment, knowledge retention, and inference efficiency.

In conclusion, while the safety mechanisms of our method introduce some initial inference overhead, the gap diminishes as batch size increases, with a manageable difference of 1.891 minutes at 10 batches. Combined with superior performance, DNAlign proves feasible for real-world deployment scenarios requiring both efficiency and quality.

\begin{figure}
    \centering
    \includegraphics[width=1\linewidth]{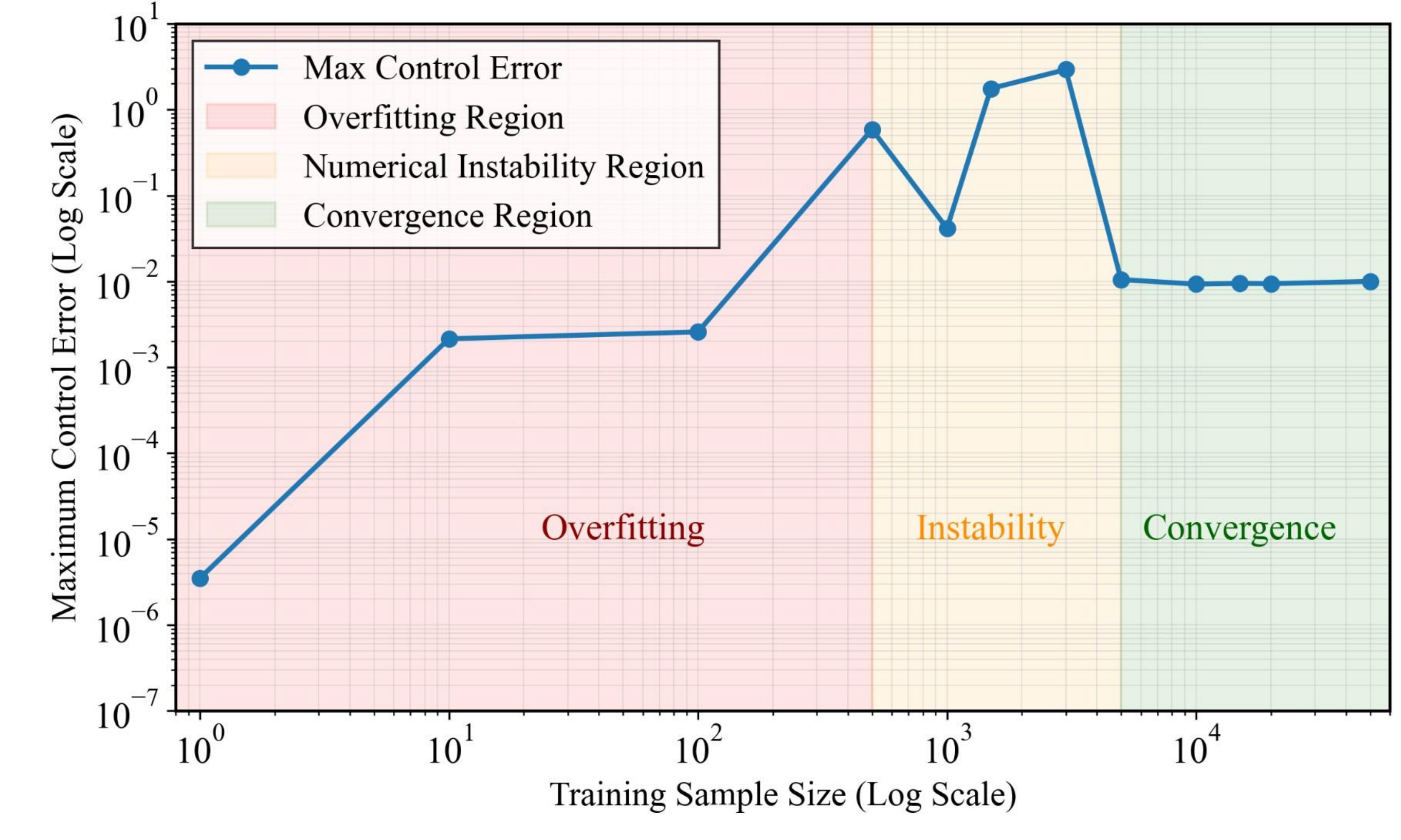}
    \caption{Research on sample parameters of projection matrix.}
    \label{fig:Sample}
\end{figure}

\subsection{Research on Sample Parameters of Projection Matrix}
To justify the design choices underlying the null-space projection module in our framework, we ablate the number of samples used to construct the projection matrix and analyze its impact on numerical stability and approximation error.
The selection of 20,000 samples for computing the projection matrix with Vicuna-7B is determined as the optimal engineering solution, supported by detailed empirical data and systematic experimental validation. As illustrated by the experimental results, the error exhibits distinct variation patterns across different sample size ranges. When the sample size is within 1 to 100, the model falls into the overfitting region, with errors maintaining at a low level (0.0000035 for 1 sample and 0.00258 for 100 samples) but lacking generalizability due to insufficient data diversity. As the sample size increases to the range of 500 to 3,000, a critical numerical instability region emerges, where the error fluctuates drastically. Specifically, the error rises to 0.587 at 500 samples, drops to 0.0415 at 1000 samples, then surges again to 1.757 at 1500 samples and reaches the peak of 2.923 at 3000 samples. This severe instability makes sample sizes in this range unsuitable for projection matrix construction. 

When the sample size exceeds 3000 and enters the convergence region (5000 to 20000 samples), the error stabilizes at a low level and shows minimal fluctuations. At 5000 samples, the error is 0.0105. At 10000 samples, it decreases slightly to 0.00929. At 15000 samples, it rises marginally to 0.00949. And at 20000 samples, it stabilizes at 0.0094. Notably, the 20,000 sample solution not only achieves a stable and low error (0.0094) but also performs comparably to the 50,000 sample variant (error 0.0100) while significantly reducing computational costs. Additionally, the 20,000 sample size far exceeds the model's hidden dimension (4096), ensuring the projection matrix is built on a complete orthogonal basis that meets the full-rank condition, thus guaranteeing mathematical rigor and excellent generalization ability. This selection also aligns with the sample size used in earlier experiments on Qianwen-7B, enabling fair cross-model comparisons.

In summary, the 20,000 sample approach achieves an optimal balance across four key dimensions: reliability (avoiding the numerical instability region), efficiency (reducing computational costs), performance (maintaining stable low errors in the convergence region), and scientific rigor (supporting cross-model comparisons). It is therefore confirmed as a fully validated and ideal solution for projection matrix computation with Vicuna-7B.

\begin{figure}
    \centering
    \includegraphics[width=1\linewidth]{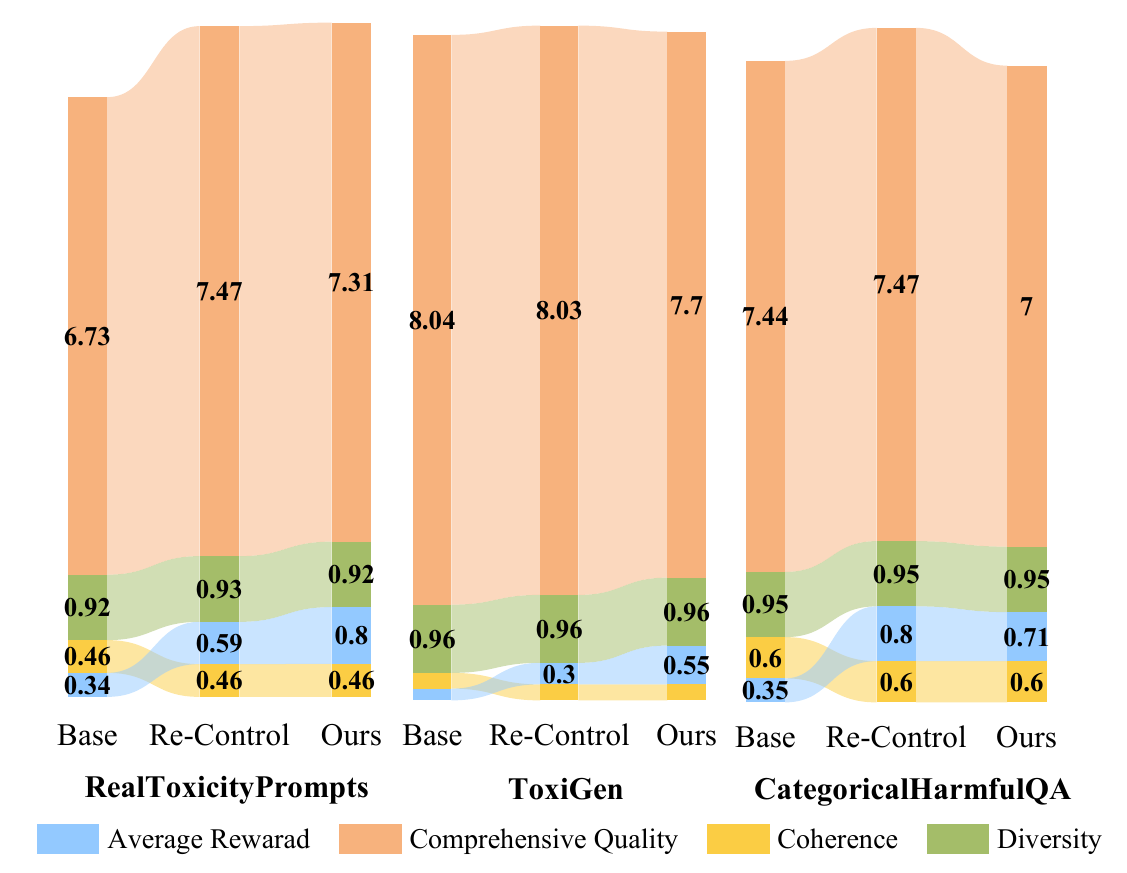}
    \caption{Generalization Performance of DNAlign on Harmful Datasets.
        Comparison of DNAlign against the base model (Qwen-7B) and RE-Control across three distinct harmful datasets: RealToxicityPrompts, ToxiGen, and Categorical-HarmfulQA. DNAlign achieves superior performance in safety alignment (average reward), notably increasing it by 137.1\% on RealToxicityPrompts and 229.4\% on ToxiGen compared to the base model. The results confirm DNAlign's strong generalization ability, effectively balancing significant safety improvements with competitive diversity and coherence across diverse harmful content distributions.}
    \label{fig:Generalization Testing}
\end{figure}

\begin{figure*}
    \centering
    \includegraphics[width=1\textwidth]{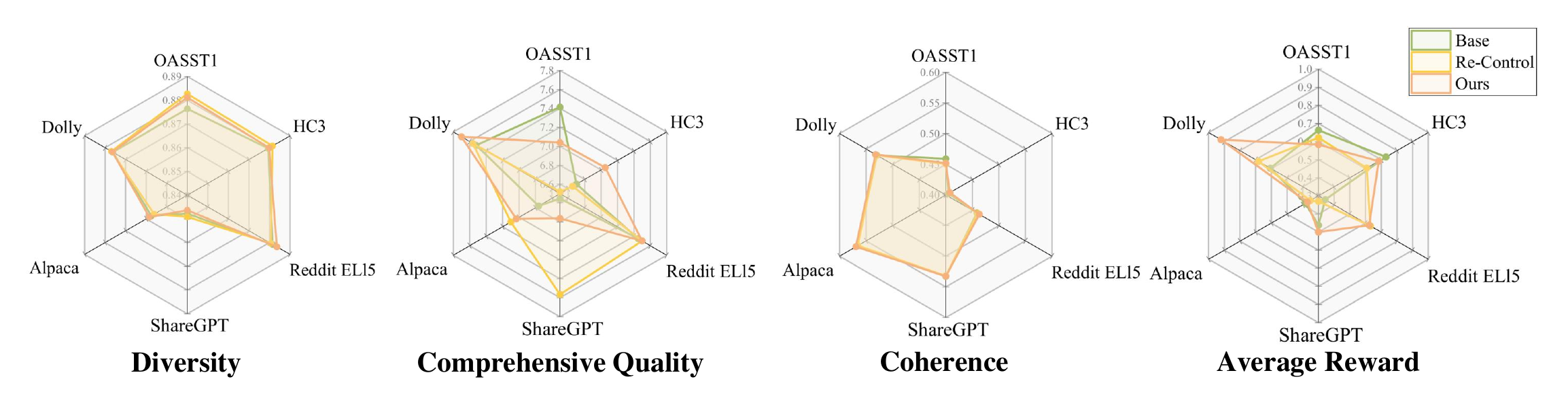}
    \caption{Performance on Harmless Datasets and Core Capability Preservation. Evaluation of DNAlign compared to the base model and RE-Control across six diverse harmless datasets, assessing core LLM capabilities. DNAlign successfully maintains competitive diversity (0.8464 to 0.8836) and coherence (0.4077 to 0.5692) scores close to the base model, demonstrating minimal degradation of core capabilities. Simultaneously, DNAlign achieves a superior average reward, outperforming RE-Control and often showing significant safety improvement (up to 81.9\% gain in average reward over the base model on specific datasets). This confirms that the null-space projection effectively balances safety alignment and utility.}
    \label{fig:Harmless}
\end{figure*}

\subsection{Generalization Testing}
The critical third research question, RQ3, mandates the evaluation of the proposed DNAlign methodology's generalization capability, specifically, its performance when facing harmful prompts that lie outside the scope of its original training distribution. This assessment is crucial for validating the robustness required for real-world deployment. To this end, we benchmark DNAlign against the base model (Qwen-7B) and the RE-Control baseline across three distinct external harmful datasets: RealToxicityPrompts, ToxiGen, and CategoricalHarmfulQA. These datasets represent diverse facets of safety risks, ranging from general toxicity to highly specific categorical harms. Our evaluation employs four rigorous metrics: diversity, coherence, comprehensive quality, and, critically, average reward for safety alignment. Across all three generalization datasets, DNAlign consistently demonstrates superior performance, particularly in fulfilling the stringent requirements of RQ3. For example, on the RealToxicityPrompts benchmark, which tests responses to model-generated toxic seeds, DNAlign achieves an average reward of 0.7999. This represents a dramatic improvement, showcasing a 137.1\% increase over the base Qwen-7B model's score of 0.3374. Importantly, this substantial safety gain is realized while maintaining highly competitive generative metrics, with diversity at 0.9241 and coherence at 0.4648, thereby mitigating the typical trade-off between safety and utility.

The effectiveness of DNAlign is further substantiated when tested against the extreme toxicity profiles in the ToxiGen dataset. Here, DNAlign’s average reward rises to 0.5465, an extraordinary 229.4\% increase relative to the base model. This significant leap underscores the method's exceptional capability in neutralizing high-risk, targeted harmful content. Even in this high-toxicity context, the fundamental quality of the generated text remains robust, evidenced by a coherence score of 0.2251 and a high diversity score of 0.9615. While acknowledging that DNAlign’s comprehensive quality score may be marginally lower than that of the unedited backbone on some specific datasets, a minor consequence of its stringent safety filtering, the overwhelming and consistent gains observed in the safety alignment metric (average reward) confirm its superior overall contribution. These collective results confirm that the DNAlign approach successfully maintains strong and reliable generalization across a wide spectrum of diverse harmful content. By demonstrating enhanced safety alignment without compromising the model's core generative capabilities, DNAlign proves its robustness and its capacity to successfully balance the dual requirements of safety and utility, making it exceptionally well-suited for high-stakes, real-world deployment scenarios.

\subsection{Impact Analysis on Harmless Datasets}

To complete our comprehensive analysis of $\text{RQ}_2$, we rigorously investigated the effect of DNAlign on purely harmless datasets. This is essential to confirm that the safety intervention does not degrade core capabilities necessary for benign tasks. Experiments were conducted on six diverse harmless datasets, comparing DNAlign against the base model (Qwen-7B) and the RE-Control baseline using four core metrics: diversity, coherence, average reward, and comprehensive quality. For experimental consistency, DNAlign was optimized with a step size of $0.2$ and $100$ gradient ascent iterations. The proposed method consistently performed at a high level across all six benign datasets, crucially maintaining performance metrics that were closely competitive with the unedited base model, while simultaneously surpassing RE-Control in key utility and safety metrics. Notably, DNAlign's diversity scores remained robust, ranging from $0.8464$ to $0.8836$, and coherence scores spanned $0.4077$ to $0.5692$, reflecting strong, preserved output consistency. 

Furthermore, DNAlign’s average reward (varying from $0.3706$ to $0.9210$) showed significant improvements over the base model and consistently outperformed the $\text{RE-Control}$ method in the vast majority of scenarios.While acknowledging that DNAlign’s comprehensive quality scores were marginally lower than the base model in some instances—a minor consequence of its stringent safety filtering—it reliably and consistently outperformed the $\text{RE-Control}$ baseline, demonstrating its superior ability to enhance safety without sacrificing core capabilities. For example, in utility-focused datasets such as Reddit ELI5 and Dolly, DNAlign exhibited superior average reward (achieving up to 81.9 improvement over the base model), all while sustaining high coherence and diversity scores. Overall, the observed stability of our method across all core metrics confirms the effectiveness of its underlying null-space projection mechanism. This mechanism successfully isolates safety-related interventions to specific subspaces orthogonal to the model’s neutral knowledge and utility functions, ensuring minimal disruption to the core capabilities. Consequently, DNAlign successfully strikes an effective and reliable balance between stringent safety alignment ($\text{RQ1}$) and core utility preservation ($\text{RQ2}$), positioning it as a robust solution for safe and effective $\text{LLM}$ deployment in real-world applications.

\subsection{Summary and Key Findings}
The comprehensive experiments confirm DNAlign's efficacy and robustness across all four research questions ($\text{RQ1}$-$\text{RQ4}$). For Safety Alignment ($\text{RQ1}$), DNAlign consistently achieved the highest Average Reward and Comprehensive Quality on the $\text{HH-RLHF}$ benchmark (e.g., $1.2258$ with $\text{Vicuna-7B}$), surpassing all baselines. For Utility Preservation ($\text{RQ2}$), $\text{Diversity}$ and $\text{Coherence}$ remained competitive with the base model on all harmless datasets, validating the null-space projection mechanism and demonstrating superior $\text{Average Reward}$ (up to $81.9\%$ gain) over baselines in benign scenarios.Crucially, DNAlign exhibited strong Generalization ($\text{RQ3}$), achieving an extraordinary 229.4 increase in Average Reward over the base model on the high-toxicity $\text{ToxiGen}$ dataset. Addressing Practical Efficiency ($\text{RQ4}$), the minimal inference overhead became negligible at larger batch sizes. In conclusion, DNAlign provides a robust and reliable framework that effectively balances stringent safety alignment with high-fidelity utility preservation, positioning it as an ideal solution for safety-critical $\text{LLM}$ deployments.

\section{Conclusion}
DNAlign is a framework that effectively balances LLM safety alignment with core knowledge preservation, addressing the trade-off seen in existing methods. By integrating dynamic control theory and null-space projection, the proposed framework ensures targeted safety interventions without disrupting the model's inherent capabilities. Experimental results demonstrate the superiority of our method across multiple datasets, outperforming baseline models like RE-Control in key metrics such as average reward, coherence, and comprehensive quality. Taken together, these findings provide affirmative answers to our four guiding research questions on safety, utility preservation, generalization, and practicality, positioning our framework as a promising solution for reliable and high-quality LLM deployment.

\section*{Author Biography}
\begin{IEEEbiography}[{\includegraphics[width=1in,height=1.25in,clip,keepaspectratio]{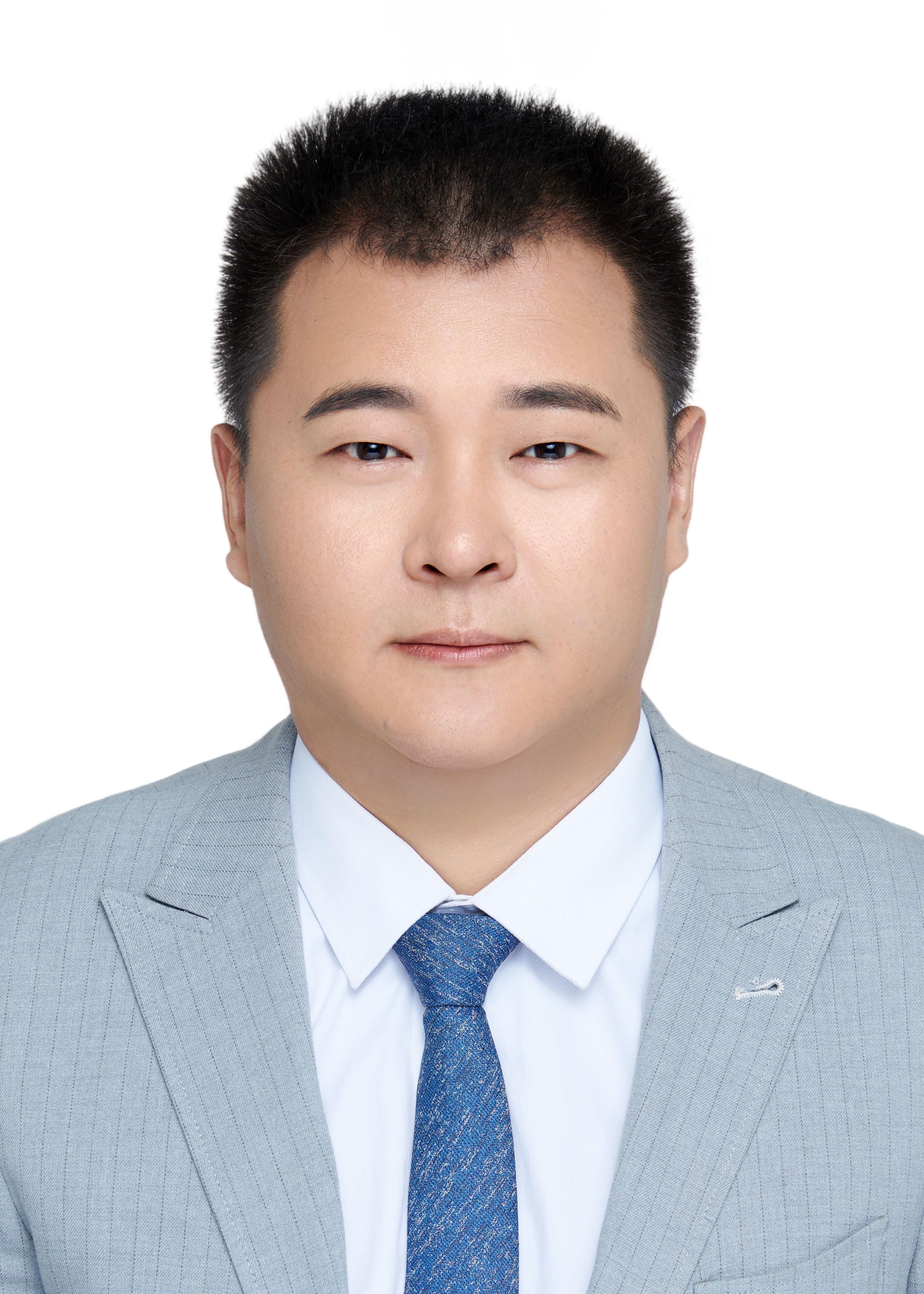}}]{Jisheng Dang}
received the Ph.D. degree from Sun Yat-sen University, China, in 2025, advised by Prof. Jianhuang Lai and Prof. Huicheng Zheng. He worked as a research fellow at the NExT++ laboratory of the National University of Singapore, advised by Prof. Tat-Seng Chua. He is now a tenured associate professor at the School of Information Science and Engineering, Lanzhou University. His research interests include multimodal learning, video understanding, and embodied intelligence. He has published several papers as the first author in major journals and conferences including IEEE TIP, TNNLS, TITS, IJCAI, and AAAI.
\end{IEEEbiography}

\begin{IEEEbiography}[{\includegraphics[width=1in,height=1.25in,clip,keepaspectratio]{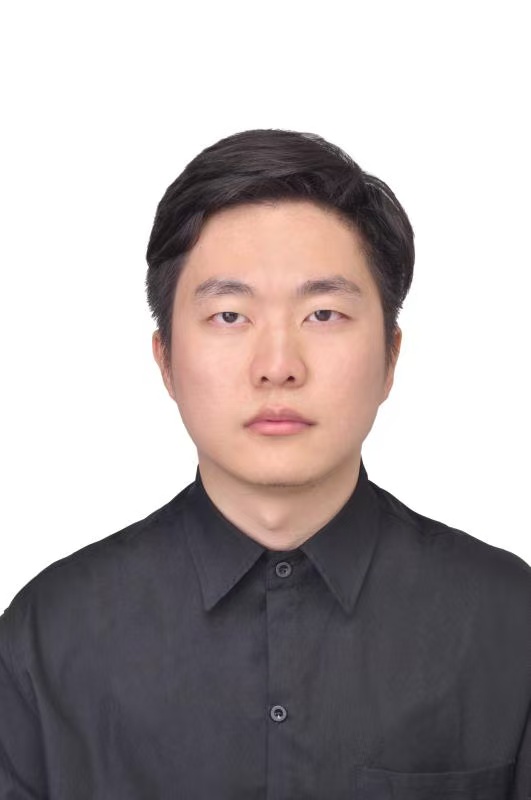}}]{Junfeng Fang}
is currently a postdoctoral researcher at the National University of Singapore (NUS). His research focuses on trustworthy large language models, including LLM safety and LLM-based agent systems. He has published over 30 papers in top-tier conferences and journals such as IEEE TPAMI, ICLR, NeurIPS, and ICML. He can be contacted at fangjf@nus.edu.sg.
\end{IEEEbiography}

\begin{IEEEbiography}[{\includegraphics[width=1in,height=1.25in,clip,keepaspectratio]{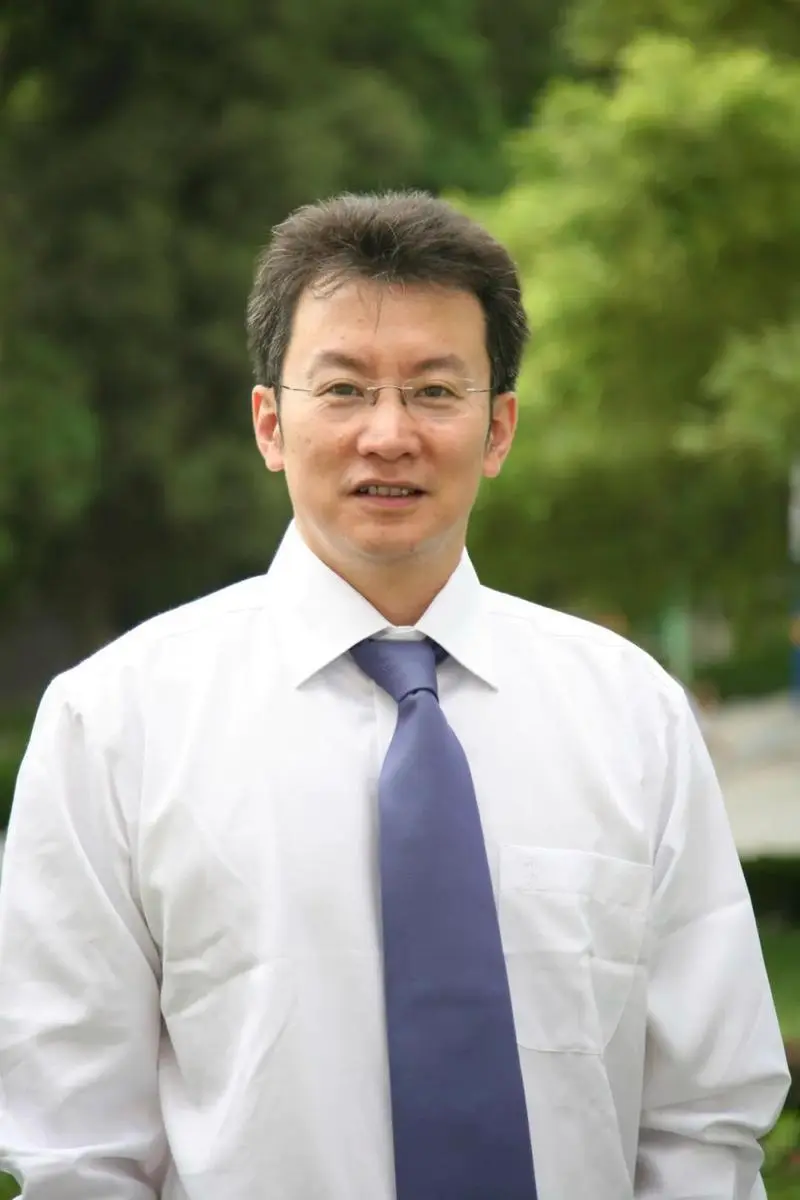}}]{Bin Hu}
(Fellow, IEEE) received the PhD degree in computer science from the Institute of Computing Technology, Chinese Academy of Science, China, in 1998. Since 2008, he has been a professor and the dean of the School of Information Science and Engineering, Lanzhou University, China. He is a Fellow of IET and AAIA, and Chair of Technical Committee on Computational Psychophysiology, IEEE SMC.
\end{IEEEbiography}

\begin{IEEEbiography}[{\includegraphics[width=1in,height=1.25in,clip,keepaspectratio]{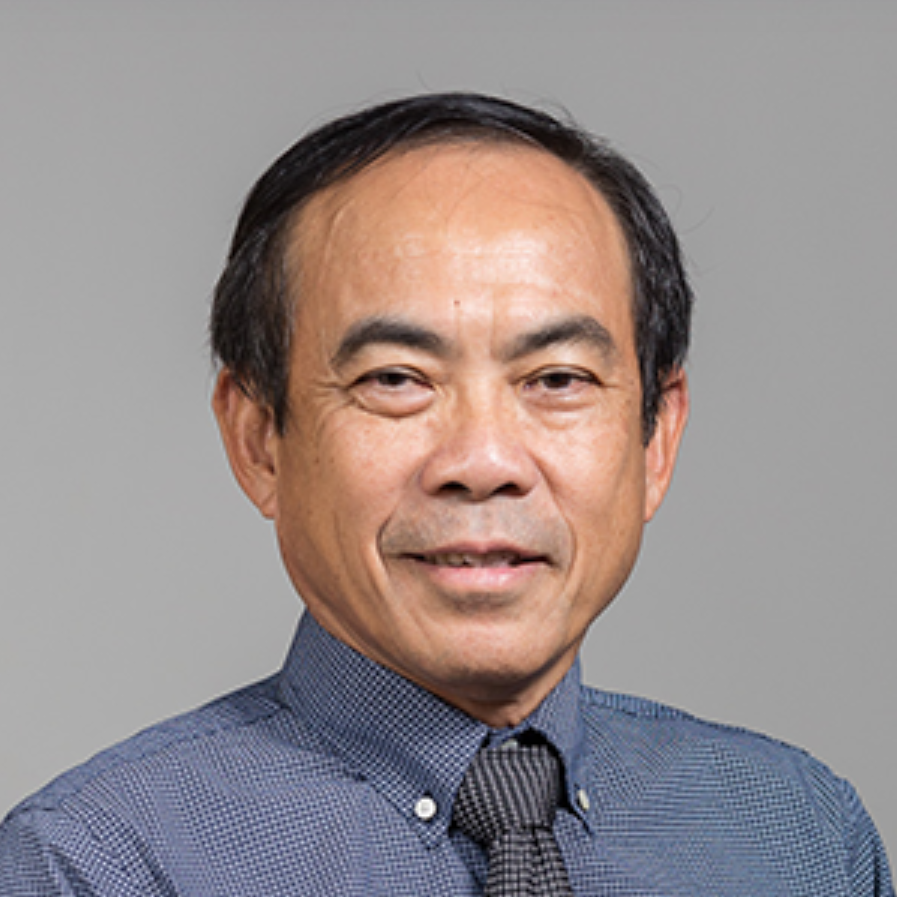}}]{Tat-Seng Chua}
received the Ph.D. degree from the University of Leeds, U.K. He is the KITHCT chair professor with the School of Computing, National University of Singapore. He is the co-director of NExT, a joint center between NUS and Tsinghua University. He is the 2015 winner of the prestigious ACM SIGMM Award and a Fellow of the Singapore Academy of Sciences. He is the co-founder of two technology startups in Singapore and a Fellow of the Singapore Academy of Sciences, with   106,330 citations on Google Scholar.
\end{IEEEbiography}

\vfill

\end{document}